\documentclass[10pt,journal]{IEEEtran}
\usepackage{lineno}
\usepackage{lineno}
\usepackage{url}
\usepackage{graphicx}
\usepackage{epstopdf}
\usepackage{subfigure}
\usepackage{amsmath}
\usepackage{comment}
\usepackage{array}
\usepackage{pbox}
\usepackage{multirow}
\usepackage{todonotes}
\usepackage{floatrow}
\usepackage{bm,graphicx}
\usepackage{amsmath}
\usepackage[inline, shortlabels]{enumitem}

\usepackage{booktabs} 
\usepackage{soul}
\usepackage{color, colortbl}
\usepackage{mathtools}
\usepackage{lscape}
\usepackage{rotating}
\definecolor{Gray}{gray}{0.9}
\definecolor{darkgray}{rgb}{0.66, 0.66, 0.66}

\begin{document}
\title{PORTRAIT: a hybrid aPproach tO cReate extractive ground-TRuth summAry for dIsaster evenT}

\author{\IEEEauthorblockN{Piyush Kumar Garg\IEEEauthorrefmark{1}, Roshni Chakraborty\IEEEauthorrefmark{2}, Sourav Kumar Dandapat\IEEEauthorrefmark{1}
}\\
	\IEEEauthorblockA{\IEEEauthorrefmark{1}Deprtment of Computer Science and Engineering\\
	Indian Institute of Technology Patna, India\\
    \IEEEauthorblockA{\IEEEauthorrefmark{2}Institute of Computer Science\\
    University of Tartu, Estonia\\
		Email: \IEEEauthorrefmark{1}piyush\_2021cs05@iitp.ac.in,
\IEEEauthorrefmark{2}roshni.chakraborty@ut.ee,
\IEEEauthorrefmark{1}sourav@iitp.ac.in
}}
	}

\IEEEtitleabstractindextext{%

\maketitle

\begin{abstract}

Disaster summarization approaches provide an overview of the important information posted during disaster events on social media platforms, such as, Twitter. However, the type of information posted significantly varies across disasters depending on several factors like the location, type, severity, etc. Verification of the effectiveness of disaster summarization approaches still suffer due to the lack of availability of good spectrum of datasets along with the ground-truth summary. Existing approaches for ground-truth summary generation (ground-truth for extractive summarization) relies on the wisdom and intuition of the annotators. Annotators are provided with a complete set of input tweets from which a subset of tweets is selected by the annotators for the summary. This process requires immense human effort and significant time. Additionally, this intuition-based selection of the tweets might lead to a high variance in summaries generated across annotators.  Therefore, to handle these challenges, we propose a hybrid (semi-automated) approach (PORTRAIT) where we partly automate the ground-truth summary generation procedure. This approach reduces the effort and time of the annotators while ensuring the quality of the created ground-truth summary. We validate the effectiveness of PORTRAIT on $5$ disaster events through quantitative and qualitative comparisons of ground-truth summaries generated by existing intuitive approaches, a semi-automated approach, and PORTRAIT. We prepare and release the ground-truth summaries for $5$ disaster events which consist of both natural and man-made disaster events belonging to $4$ different countries. Finally, we provide a study about the performance of various state-of-the-art summarization approaches on the ground-truth summaries generated by PORTRAIT using ROUGE-N F1-scores.
\end{abstract}

}
\maketitle

\IEEEdisplaynontitleabstractindextext

\IEEEpeerreviewmaketitle

\section{Introduction} \label{s:intro}
\par Social media platforms, such as Twitter, are important mediums where users share information during disaster events~\cite{imran2016twitter}. People from the affected locations share messages about their urgent needs while government organizations, volunteers and humanitarian agencies share information about the availability of resources and services. Government agencies utilize these information from the affected locations to ensure immediate relief operations~\cite{castillo2016big}. Several research works have highlighted the role of social media websites, such as Twitter, for effective disaster management~\cite{basu2019extracting, dutt2019utilizing, ghosh2018exploitation}. However, tweets are inherently short and comprise of grammatical errors, abbreviations and informal language, making it highly challenging to identify the relevant information. Additionally, the huge volume of these messages increases the challenges for government organizations, humanitarian agencies, and volunteers to identify relevant information manually~\cite{alam2020descriptive, Alam2021humaid}.

\begin{table*}[ht]
    \caption{We show the details of available $6$ disaster datasets, including dataset name, number of tweets, summary length, country, continent, and disaster type.}
    \label{table:existdata}
    \resizebox{\textwidth}{!}{\begin{tabular} {lccccc}
        \toprule
        {\bf Dataset name} & {\bf Number of} & {\bf Summary} & {\bf Country} & {\bf Continent} & {\bf Disaster type}\\ 
        & {\bf tweets} & {\bf length} & & & \\ \midrule
        
        \textit{Sandy Hook Elementary School Shooting}  & 2080 & 36 tweets & United States of America  & USA & Man-made   \\ 
        \textit{Uttrakhand Flood}                       & 2069 & 34 tweets & India          & Asia & Natural  \\ 
        \textit{Hagupit Typhoon}                        & 1461 & 41 tweets & Philippines    & Asia & Natural  \\ 
        \textit{Hyderabad Blast}                        & 1413 & 33 tweets & India          & Asia & Man-made \\ 
        \textit{Harda Twin Train Derailment}            & 4171 & 250 words & India          & Asia & Man-made \\ 
        \textit{Nepal Earthquake}                       & 5000 & 250 words & Nepal          & Asia & Natural  \\ \bottomrule
    \end{tabular}}
\end{table*}

\par To mitigate these issues, recent research works~\cite{garg2022ontodsumm, dusart2021tssubert, rudra2019summarizing, dutta2018ensemble, saini2022microblog} have proposed automated tweet summarization approaches which can handle the huge number of user tweets posted during a disaster event. Summary generated by these approaches can aid government agencies to identify important information, such as identification of the required resources across affected locations, infrastructural damage, etc. However, it is noticed that the quality of the summary produced by existing approaches varies significantly across different disaster datasets. This is mainly because of the high variance across different datasets in terms of the location, type and severity of disasters. Furthermore, due to the lack of ground-truth summary, existing algorithms can not be thoroughly tested for robustness. To check the effectiveness and robustness of a summarization approach, we require a good number of ground-truth summaries of disaster events from different locations and of different types. Although \cite{dutta2018ensemble} and \cite{rudra2018extracting} have provided ground-truth summary of $6$ datasets (shown in Table~\ref{table:existdata}) which is of huge help to the research community, it is not sufficient for testing. Although addition of new datasets will surely improve this scenario, ground-truth summary generation is a costly task in terms of time and manual effort. This scenario motivates us to come up with a strategy which can reduce human effort and time.

\par A good summary of an event must capture the relevant and diverse aspects of the event as well as it should cover all the important aspects/topics~\footnote{From now onward, we refer to aspect and topic both by topic in this paper.} of the event. So to come up with a good ground-truth summary, an annotator requires initially to identify the topics of each tweet, followed by determination of the relative importance of each topic with respect to the other topics and finally, select tweets from different topics based on the importance of a tweet in its own topic and the importance of topic respect to the event for the final summary. This process requires extensive manual efforts and significant amount of time from the annotators. Moreover, the quality of the final summary depends on the wisdom and understanding of the annotator as all the intermediate steps followed by annotators are subjective. Therefore, we can not rely on a summary generated by a single annotator~\cite{poddar2022caves, rudra2018extracting, dutta2018ensemble}. Existing approaches suggest that we should have at least $3$ annotators to generate $3$ different summaries. Evaluation of an automatically generated summary should be compared with each of the individual summaries, and an average score of the comparison results across the individual summaries to ensure that the proposed summary is consistent and fair.

\par Although there are several 
existing research works~\cite{pasquali2021tls, dusart2021issumset, he2020tweetsum, dusart2020capitalizing, feigenblat2021tweetsumm} which create ground-truth summary of an event, only a few existing research works~\cite{nguyen2018tsix, cao2016tgsum} discuss guidelines/approaches how to generate a ground-truth summary. These existing works can further be segregated on their proposed approach into fully-automated approach~\cite{cao2016tgsum} for ground-truth creation of news multi-document summarization dataset guided by tweets, semi-automated approach~\cite{nguyen2018tsix} for ground-truth creation of Twitter social events, and completely manual approach~\cite{feigenblat2021tweetsumm, he2020tweetsum, antognini2020gamewikisum, nguyen2016vsolscsum}. However, fully automated ground-truth creation method~\cite{cao2016tgsum} is practically a summarization approach without any human intervention. Therefore, there is no justified reason to treat the created summary as ground-truth. In the semi-automated method~\cite{nguyen2018tsix}, the authors used a number of summarization methods to select a subset of tweets for annotators. There are few practical issues in this approach as i) it relies on a specific set of summarization algorithms which might result good for a specific dataset and bad for some other dataset ii) it identifies topics by unsupervised clustering methods which suffer from vocabulary overlap issue~\cite{garg2022ontodsumm}. Moreover, these existing ground-truth creation guidelines/approaches are not directly applicable to ground-truth summary creation for disaster events. This is mainly due to non-fulfilment of the summary objectives, high vocabulary overlap across clusters in fully-automated and semi-automated approaches, domain-dependent annotation instructions, and high variance in generated summaries across annotators. There are a few existing disaster summarization approaches~\cite{dutta2018ensemble, rudra2018extracting, rudra2015extracting} which provide the ground-truth summary. However, in the above-mentioned approaches, ground-truth summary is generated based on the wisdom and intuition of the annotators, where the annotators are provided with all the tweets with respect to the disaster, and he/she has to select the tweets manually.

\par In this paper, we propose a hybrid (semi-automated) approach (PORTRAIT) to generate the ground-truth summary where we automate the process partly (without compromising the quality of ground-truth) so that the annotator's efforts are reduced. Along with that, we provide guidelines to ensure consistent summaries. Therefore, we propose a systematic semi-automated approach for ground-truth summary generation. We validate the effectiveness of PORTRAIT on $5$ disaster events by comparing ground-truth summary generated by PORTRAIT with ground-truth summary generated by existing approaches. We perform both qualitative and quantitative comparisons on the three most important characteristics of summary, namely \textit{coverage, relevance, and diversity}~\cite{chakraborty2019tweet,chakraborty2017network}. We perform qualitative comparison with the help of $3$ meta annotators who rated both the summaries for \textit{coverage, relevance, and diversity} and utilize metrics to capture \textit{coverage, relevance, and diversity} for qualitative as well as quantitative comparison. Using both qualitative and quantitative comparisons, it is confirmed that the quality of ground-truth summary generated by PORTRAIT is better compared to the summary generated by annotators' intuition as well as ground-truth summary generated by the existing semi-automated approach. 
Additionally, we release the ground-truth summaries for $5$ disaster events, which belong to different types and from different countries, such as the United States of America, Haiti, Mexico, and Pakistan. Our major contributions can be summarized as follows:

\begin{enumerate}

    \item We propose a semi-automated approach (PORTRAIT) to generate the ground-truth summary for disaster events. PORTRAIT reduces the effort and time of annotators.
   
    \item We provide quantitative and qualitative analysis of the effectiveness of PORTRAIT in ground-truth summary generation. Comparison result confirms that PORTRAIT ensures quality ground-truth summary.
    
    \item We prepare and release the ground-truth summary for $5$ disaster datasets of different locations and types, which would be highly helpful for the research community.
    
    \item To verify the quality of generated ground-truth summary by PORTRAIT, we have added two additional fields, namely \textit{relevance label} and \textit{explanations}. \textit{Relevance label} is a categorical variable which can take values as \textit{high}, \textit{medium} or \textit{low} and \textit{explanation} provides the possible reasoning behind the \textit{relevance label}. We provide this information for $5$ datasets which we release.
    
   \item We also compare  $13$ existing summarization approaches on these datasets, which might help the research community in understanding the performance of existing summarization algorithms.

\end{enumerate}

\par The rest of the paper is organized as follows. We discuss related works in Section~\ref{s:rworks}. In Section~\ref{s:data}, we provide the details of datasets and discuss the details of PORTRAIT in Section~\ref{s:prop}. In Section~\ref{s:res}, we discuss results where we provide the qualitative and quantitative comparison results of PORTRAIT summary in Section~\ref{s:analy} and Section~\ref{s:justifi}, respectively. We discuss the experiment details, and results for performance comparison of the existing summarization approaches on the ground-truth summaries generated by PORTRAIT in Section~\ref{s:exp}. Finally, we conclude the paper in Section~\ref{s:con}.  

\section{Related Works} \label{s:rworks}

\par Summarization provides a comprehensive gist which includes all the important aspects of an event. This becomes very important when event comprises of sufficiently large amount of text/tweets where there is high chance of duplicate information and noise. This attracts a large group of researchers, and we find a very rich literature on summarization work for different event types.

\par Tweet summarization approaches proposed for disaster events can be broadly categorized in terms of methodology as content and context-based approaches~\cite{rudra2015extracting, li2021twitter}, graph-based approaches~\cite{dutta2015graph} and deep learning-based approaches~\cite{de2019time}. However, irrespective of the approach, any disaster tweet summarization approach requires a good number of ground-truth summaries of different disaster events from different locations and types for the testing of robustness. There is an important point to be noted that disaster datasets collected from different locations and of different types exhibit a high variance~\cite{garg2022ontodsumm}. Hence, it is quite likely a proposed summarization algorithm might be suitable for a set of input datasets while not appropriate for different sets of inputs. Till date, we found a very limited ground-truth dataset for disaster event and hence there is an immediate need to create adequate amount of ground-truth summary of disaster events from different locations and of different types. However, generation of ground-truth summaries for disaster events has several challenges, and very few disaster summarization approaches discuss the procedure to generate the ground-truth summary. Therefore, we initially discuss existing literature for ground-truth summary generation for different applications, such as multiple documents, customer-agent interaction, social media interactions, etc., which can provide us with critical insights on how to develop ground-truth summary generation algorithms. We, finally, discuss the ground-truth summary generation for tweets related to disaster events specifically.

\par Existing ground-truth summary generation approaches for different applications are either extractive~\cite{feigenblat2021tweetsumm, yela2021multihumes, pasquali2021tls, nguyen2016solscsum} or abstractive~\cite{poddar2022caves, chen2022summscreen, chen2021dialogsum, fabbri2019multi, yasunaga2019scisummnet}. Existing extractive ground-truth summary generation approaches can be further categorized as automated approaches~\cite{cao2016tgsum}, semi-automated approaches~\cite{nguyen2018tsix}, or manual annotation-based approaches~\cite{feigenblat2021tweetsumm, he2020tweetsum, antognini2020gamewikisum, nguyen2016vsolscsum} whereas abstractive summarization approaches found in the literature are only manual annotation based approaches~\cite{poddar2022caves, wang2022clidsum, chen2022summscreen, chen2021dialogsum, narayan2018don}. Manual annotation-based approaches provide the complete set of input sentences to an annotator who selects the sentences into the summary on the basis of their wisdom and intuition. While some of these approaches provide a specific set of instructions~\cite{feigenblat2021tweetsumm, poddar2022caves, chen2021dialogsum, hasan2021xl, guy2021tweetsumm} to the annotators, the others do not provide any specific instruction~\cite{yela2021multihumes, dusart2021issumset, he2020tweetsum, nguyen2016solscsum, wang2022clidsum, yasunaga2019scisummnet, gliwa2019samsum, narayan2018don}. In case of extractive ground-truth summarization approaches without instructions~\cite{dusart2021issumset, dusart2020capitalizing,he2020tweetsum,nguyen2016vsolscsum}, ask the annotator to gauge the importance of a sentence to decide whether it should be selected into the summary, while for abstractive ground-truth summarization approaches without instructions~\cite{fabbri2019multi,narayan2018don} ask the annotator to gauge the importance of a keyword to decide whether it should be selected into the summary. However, understanding the importance of a sentence or the keyword only on the basis of intuition and wisdom can be very difficult for an annotator and further, can lead to inconsistent summaries across annotators. To handle this challenge, few existing manual annotation-based ground-truth summary generation approaches provide more detailed guidelines to help the decision-making of the annotators, such as examples of informative and uninformative summaries~\cite{feigenblat2021tweetsumm}, description of the summary objectives, like, coherence, readability, abstractivity, coverage, and diversity~\cite{poddar2022caves} or specific instruction related to the application, such as understanding of the customer requirements and the desired agent response~\cite{guy2021tweetsumm}. Although these guidelines are immensely helpful for the annotators, none of these guidelines intends to reduce the effort of the annotators. Additionally, since all of these guidelines are subjective and generic, they can not ensure consistency across annotators, and therefore, the summary generated by different annotators might vary.

\begin{table*}[ht]
    \caption{We show the details of $5$ disaster datasets for which we create the summary, including dataset number, dataset name, number of tweets, number of tweets after category classification (which we will discussed in detail in Section~\ref{s:ch}), summary length, country, continent, and disaster type.} 
    \label{table:propdata}
    \centering
    \resizebox{\textwidth}{!}{\begin{tabular} {clcccccc}
        \toprule
        {\bf Num} & {\bf Dataset name} & {\bf Number of} & {\bf Number of tweets} & {\bf Summary} & {\bf Country} & {\bf Continent} & {\bf Disaster type}\\ 
        
        & & {\bf tweets} & {\bf after category} & {\bf length} & & & \\
        & & & {\bf classification} & & & & \\\midrule
        
        $D_1$ & \textit{Los Angeles International Airport Shooting} & 1409 & 935 & 40 tweets & United States of America  & USA   & Man-made \\ 
        $D_2$ & \textit{Hurricane Matthew}                          & 1654 & 1477 & 40 tweets & Haiti          & USA   & Natural  \\ 
        $D_3$ & \textit{Puebla Mexico Earthquake}                   & 2015 & 1896 & 40 tweets & Mexico         & USA   & Natural  \\ 
        $D_4$ & \textit{Pakistan Earthquake}                        & 1958 & 1781 & 40 tweets & Pakistan       & Asia & Natural  \\ 
        $D_5$ & \textit{Midwestern U.S. Floods}                     & 1880 & 1575 & 40 tweets & United States of America  & USA   & Man-made \\ \bottomrule
    \end{tabular}}
\end{table*}

\par To reduce human effort and inconsistency across ground-truth summaries generated by different annotators, several existing approaches have proposed automated or semi-automated approaches in different applications. For example, Cao et al.~\cite{cao2016tgsum} proposed an automated approach which initially segregates tweets into clusters, followed by the selection of representative tweets from each cluster by Integer Linear Programming (ILP) based optimization technique to generate the summary. Although an automated approach reduces human efforts completely, this approach does not include the human wisdom and intuition required to resolve the subjective task of ground-truth summarization. Therefore, it is only a summarization approach which can not be treated as ground-truth summary generation approach. On the basis of these existing approaches, we observe that neither automated nor manual approaches can ensure consistent ground-truth summaries with minimum human effort. In order to resolve this, Nguyen et al.~\cite{nguyen2018tsix} proposed a semi-automated approach which initially segregates the tweets into clusters on the basis of their topic. In the next step, Nguyen et al.~\cite{nguyen2018tsix} employ $3$ existing summarization algorithms such that each algorithm selects the most informative tweets from each cluster into a reference tweet set. Therefore, the reference tweet set includes all the informative tweets by $3$ summarization algorithms from all the clusters. Finally, the annotator manually selects the tweets from reference tweet set into the ground-truth summary on the basis of their wisdom and intuition. Although this approach integrates both automation and manual-based ground-truth summary generation, which reduces human effort, it has a few shortcomings. For example, identifying topics by clustering is error-prone as clustering primarily groups tweets based on vocabulary. It is found many times that the same words are being used in different contexts and meanings. Moreover, this approach relies on $3$ specific summarization approaches to select important tweets from each cluster. There is a high chance that this approach will be highly data dependent which means it might produce good results for certain datasets while it may result bad for some other datasets.

Similarly, there are several existing disaster ground-truth summary creation approaches, such as abstractive~\cite{nguyen2022rationale, nguyen2022towards, rudra2016summarize} or extractive~\cite{rudra2015extracting, dutta2018ensemble, rudra2018extracting}. To the best of our knowledge, we found that all of these approaches are manually generated ground-truth summary generation approaches where they  generate the summary without any help of instructions. As previously discussed, manual ground-truth summary generation approaches might not ensure consistency across annotators, fail to ensure objectives of summarization and require a huge amount of human effort and time. Further, generation of ground-truth summary is a subjective task, so we can not depend on only one annotator for the summary, and we require at least $3$ annotators for their individual summaries~\cite{poddar2022caves, dutta2018ensemble, rudra2018extracting}, thereby, increasing the effort and time from annotators by at least $3$ times.  Therefore, in this paper, we propose a semi-automated approach (PORTRAIT) wherein we provide a formalized set of steps to be followed to generate a summary and furthermore, we provide automated solutions to several of these steps, which reduces the annotator's effort and time and can ensure consistency across annotators. We discuss datasets details next.

\section{Datasets} \label{s:data}
\par In this Section, we discuss the disaster events for which the ground-truth summaries are available as well as the disaster events for which we prepare the ground-truth summary.

\par Dutta et al.~\cite{dutta2018ensemble} provided the ground-truth summaries for \textit{Sandy Hook Elementary School Shooting}~\footnote{https://en.wikipedia.org/wiki/Sandy\_Hook\_Elementary\_School\_shooting}, \textit{Uttarakhand Flood}~\footnote{https://en.wikipedia.org/wiki/2013\_North\_India\_floods}, \textit{Hagupit Typhoon}~\footnote{https://en.wikipedia.org/wiki/Typhoon\_Hagupit\_(2014)}, and \textit{Hyderabad Blast}~\footnote{https://en.wikipedia.org/wiki/2013\_Hyderabad\_blasts} and Rudra et al.~\cite{rudra2018extracting} provided for \textit{Harda Twin Train Derailment}~\footnote{https://en.wikipedia.org/wiki/Harda\_twin\_train\_derailment} and \textit{Nepal Earthquake}~\footnote{https://en.wikipedia.org/wiki/April\_2015\_Nepal\_earthquake}, respectively. We show the details of these $6$ disaster events in Table~\ref{table:existdata}.


\par In this paper, we propose a hybrid approach (PORTRAIT) to generate the ground-truth summary with minimum human intervention and prepare ground-truth summaries of $5$ disaster events, such as \textit{Los Angeles International Airport Shooting} $(D_1)$, \textit{Hurricane Matthew} $(D_2)$, \textit{Puebla Mexico Earthquake} $(D_3)$, \textit{Pakistan Earthquake} $(D_4)$, and \textit{Midwestern U.S. Floods}. $(D_5)$. We have taken $D_1$ and $D_2-D_5$ disaster datasets from \cite{olteanu2015expect} and \cite{Alam2021humaid}, respectively. We specifically select datasets such that it covers different types of disasters, such as natural and man-made, and different continents, such as Asia and USA. We provide the details of these disaster events in Table~\ref{table:propdata}. 

\begin{table*}[ht]
    \caption{Some example of tweets text of two disaster events, such as \textit{Hurricane Matthew} $(D_2)$ and \textit{Pakistan Earthquake} $(D_4)$.}
    \label{table:Tweets}
    \centering 
    \begin{tabular}{cp{0.7\linewidth}} \toprule
    \textbf{Disaster event} & \textbf{Tweet text} \\ \midrule
        
                                & \#Jamaica Haiti: Hurricane Matthew: 350,000 people in need of assistance, 15,623 are displaced  \#crisismanagement \\ 
                                & 5 people in Haiti died and at least 10 others injured from incidents related to Hurricane \#Matthew, per Haitis Civil Protection Service.  \\ 
    \textit{Hurricane Matthew}  & RT @B911Weather: UPDATE: Death toll from Hurricane \#Matthew climbs to at least 25, most deaths occurred in Haiti - NBC News   \\
                                & RT @winknews: 3,214 homes destroyed in Haiti by Hurricane Matthew. 350,000 estimated to need some kind of assistance. 10 people killed.  \\ 
                                & U.S. providing \$400,000 in aid to Haiti and Jamaica for Hurricane Matthew \\ \midrule    
            
                                & \#earthquake. 22 people lost life including an army soldier while 160 people got injured. Three communication bridges near Jatlan damaged. \\ 
                                & \#Earthquake in \#Pakistan: Death tolls rises to 30 with over 450 injured. We are sad over losses. Prayers for early recovery of injured and souls departed during earthquake. \\ 
    \textit{Pakistan Earthquake}& NDMA distributes rations, water bottles and tents among affected families as part of relief operation in \#Kashmir  \#Pakistan. \\ 
                                & World Health Organization @WHO hands over medicines \&amp; surgical equipment to Pakistan for \#earthquake victims of \#Mirpur. \\ 
                                & 19 dead, over 300 injured as earthquake shakes parts of Pakistan.  \\ \bottomrule
    \end{tabular}
\end{table*}

\begin{enumerate}
    \item \textit{$D_1$}: This dataset is based on the tweets related to the terrorist attack on the \textit{Los Angeles International Airport Shooting}~\footnote{https://en.wikipedia.org/wiki/2013\_Los\_Angeles\_International\_Airport\_shooting} on November, $2013$ in California in which $1$ person was killed and more than $15$ people were injured~\cite{olteanu2015expect}.  
    
    \item \textit{$D_2$}: This dataset is based on the tweets related to the devastating impact of the terrible hurricane, \textit{Hurricane Matthew}~\footnote{https://en.wikipedia.org/wiki/Hurricane\_Matthew} on October, $2016$ in Haiti which caused the death of $603$ people, around $128$ people were missing, and the estimated damage was around \$$2.8$ billion USD~\cite{Alam2021humaid}.

    \item \textit{$D_3$}: This dataset is based on the tweets related to the \textit{Puebla Mexico Earthquake}~\footnote{https://en.wikipedia.org/wiki/2017\_Puebla\_earthquake} on September, $2017$ in Mexico City in which $370$ people were dead and more than $6000$ people were injured~\cite{Alam2021humaid}.

    \item \textit{$D_4$}: This dataset is based on the tweets related to the \textit{Pakistan Earthquake}~\footnote{https://en.wikipedia.org/wiki/2019\_Kashmir\_earthquake} on September, $2019$ in which around $40$ people were dead, $850$ people were injured, and around $319$ houses were damaged~\cite{Alam2021humaid}.

    \item \textit{$D_5$}: This dataset is based on the tweets related to the \textit{Midwestern U.S. Floods}~\footnote{https://en.wikipedia.org/wiki/2019\_Midwestern\_U.S.\_floods} in which around $14$ million people were affected, and  the estimated damage was around \$$2.9$ billion USD~\cite{Alam2021humaid}. 
    
\end{enumerate}

For pre-processing, we perform conversion of cases, lemmatization, removal of URLs, stop words, white-spaces, punctuation marks, and emoticons. We remove Twitter-specific keywords~\cite{arachie2020unsupervised}, such as usernames and hashtags, as we consider only the text of the tweets. We also remove the duplicate tweets and retweets and follow Alam et al.~\cite{alam2018crisismmd} to remove noise, i.e., remove any word consisting of less than $3$ characters except disaster-specific keywords~\cite{garg2022ontodsumm}. We show the details of $D_1$-$D_5$ and gold standard summary length in Table~\ref{table:propdata} and make it publicly available \footnote{https://drive.google.com/drive/folders/15x-bfdvkvTlu7b44zrNYwUcCiFvCSmFZ?usp=sharing}. We show some examples of tweets for $D_2$ and $D_4$ in Table~\ref{table:Tweets}.

\section{Proposed Approach} \label{s:prop}
\par In this Section, we elaborate the process of hybrid ground-truth summary generation approach (PORTRAIT) along with justification about which part is automated and which part is left for the human annotators. We also provide a detailed discussion of the process adopted for annotator selection.

\subsection{Ground-truth Summary Generation} \label{s:ch}
\par To ensure a good quality summary, an annotator needs to make multiple decisions for various tasks, such as identification of the topic of each tweet, assessment of the importance of the topic with respect to the disaster event, determining the importance of a tweet with respect to the topic and finally, select or leave the tweet into the ground-truth summary on the basis of both the importance of the tweet with respect to the topic and importance of topic with respect to the disaster. These tasks either may be performed explicitly or implicitly by intuition. We observe that in all existing research works that an annotator~\cite{rudra2015extracting, dutta2018ensemble, rudra2018identifying} manually identifies the importance of each tweet with respect to the disaster event and then, decides whether it should be part of the summary or not based on intuition. These approaches mainly depend on the wisdom of the annotators to select the tweets from a flat set of tweets related to a disaster. This might lead to high variance in the summaries generated by different annotators as in every step it depends on human intuition, which varies across annotators. Moreover, it might also fail to preserve all the intended features of a good summary. Along with that, it requires extensive manual effort and time from the annotators. Therefore, we propose PORTRAIT to generate the ground-truth summary where we reduce the effort and time of the annotators by providing a sub-set of the most informative tweets from each topic. Additionally, this also can ensure consistency among the different summaries across annotators. We discuss the proposed PORTRAIT next. 

\begin{table}[!ht]
    \caption{We show the number of tweets in each topic for $4$ disaster datasets, such as $D_1$, $D_2$, $D_4$, and $D_5$.}
    \label{table:numsub}
    \centering
    \begin{tabular} {lcccc}
        \toprule
        {\bf Topic} & {\bf $D_1$} & {\bf $D_2$} & {\bf $D_4$} & {\bf $D_5$} \\ 
        \midrule 
        
        Affected Population     & 380 & 250  & 440 & 73   \\ 
        Early Warning           & 344 &  37  & 42  & 49   \\ 
        Emergency Exercises     &  8  &  51  & 12  & 24   \\ 
        Emotional Distress      &  13 &  1   & 14  & -    \\ 
        Humanitarian Event      &  9  &  8   & 7   & 5    \\ 
        Impact                  &  24 &  39  & 103 & 62   \\ 
        Infrastructure Damage   &  -  & 169  & 202 & 158  \\ 
        Volunteering Support    &  59 & 591  & 323 & 1113 \\ 
        Prayer                  &  97 & 329  & 638 & 89   \\ \bottomrule     
    \end{tabular} 
\end{table}

\par As discussed earlier, a number of steps are required to come up with the summary from a flat set of tweets which includes topic identification of each tweet, assessment of topic importance, and final selection of tweets to ensure all the important aspects are covered. From the existing literature, it is well understood that the first step of this sequential process which is topic identification can be automated with very high accuracy. There are a number of approaches that can be adopted for automated topic identification. We have chosen Garg et al.~\cite{garg2022ontodsumm} to automatically identify the category/topic of a tweet as it was specially designed for disaster tweet classification based on disaster ontology and reported very high F1-score ($0.98$) as classification accuracy by considering only those tweets which could be classified using this approach. Our observations indicate that the tweets which are not classified by Garg et al.~\cite{garg2022ontodsumm} are either irrelevant or comprise of very less information. Therefore, for our next task, we do not consider the tweets whose category could not be determined using automated method. We show the number of tweets which we classified using automated method in Table~\ref{table:propdata}. The next task in the sequential process for PORTRAIT is the assessment of the relative importance of each topic with respect to the disaster event. We find that the relative importance of topics with respect to corresponding disaster event varies significantly across disasters~\cite{garg2022ontodsumm}, and identifying it automatically could be highly error-prone. So, we believe this task should be performed by human annotators to ensure high-quality ground-truth summary. In the sequential process of annotation, understanding the importance of a tweet with respect to the topic could be considered as the next task. However, this becomes highly time-consuming for the annotators if a topic consists of a huge number of tweets. For example, the number of tweets that belong to different topics, such as \textit{Volunteering Support} and \textit{Affected Population} are $1113$ and $440$ in \textit{Midwestern U.S. Floods}~\footnote{https://en.wikipedia.org/wiki/2019\_Midwestern\_U.S.\_fl.oods} and \textit{Pakistan Earthquake}~\footnote{https://en.wikipedia.org/wiki/2019\_Kashmir\_earthquake}, respectively, as shown in Table~\ref{table:numsub}. So, to reduce the efforts of an annotator, we provide only a subset of highly ranked tweets (on the basis of informativeness) from all the tweets that belong to that topic. As highly ranked tweets are more likely to be selected into the summary. Although there are a number of existing approaches for ranking  tweets~\cite{garg2022entropy, nguyen2022towards, rudra2019summarizing, rudra2015extracting}, we adopted Disaster specific Maximal Marginal Relevance (DMMR)~\cite{garg2022ontodsumm}. We choose DMMR over other approaches, as it considers the specific information of each topic related to disaster events and has been proven to be the most effective for disaster events. We use this automated ranking for selection only if the number of tweets in a topic/category is more than $25$. For the topic with more than $25$ tweets, we select the top $25\%$ most informative tweets by DMMR. However, if the number of tweets in the top $25\%$ is less than $25$, then we keep top $25$ tweets based on DMMR score. We provide the selected tweets finally to the annotators. By this automatic selection of the most informative tweets by DMMR from each topic, we reduce the number of tweets to be read by the annotators significantly, and an annotator only reads around $26.37-30.59\%$ of the classified tweets for $D_1$ to $D_5$ dataset. Additionally, as we select  a significant percentage of tweets from each topic, it is most unlikely that we will lose any important tweet which was supposed to be part of the summary. We experimentally validate this in Section~\ref{s:justifi}. However, we do not provide any associated DMMR score for the selected tweets when we provide it to the annotators as it might be misleading. Finally, we rely on an annotator to select the set of tweets from each topic into the summary as this is a highly subjective task. We provide annotators with a set of instructions/guidelines to help them.

\begin{enumerate}
    \item Annotators are instructed to read about the disaster event from external and trusted sources of information. 
    \item Annotators are also instructed to go through a set of example tweets and corresponding topic descriptions created by us. This is done for all the topics. An overview of this information is shown in Table~\ref{table:Tsubthem}.
    \item Annotators need to select the tweets from each topic on the basis of wisdom and intuition. The annotator must consider the importance of the topic with respect to the disaster and the importance of a tweet with respect to the topic to decide whether a tweet should be selected or not. An annotator can even decide not to select any tweet from a topic if he/she feels the topic/tweets of that topic/category is not important for the disaster event.
\end{enumerate}

\begin{table*}[!ht]
    \caption{We show a snippet of descriptions of different topics along with an example tweet.}
    \label{table:Tsubthem}
    \resizebox{\textwidth}{!}{\begin{tabular} {p{0.55\linewidth}p{0.45\linewidth}}
        \toprule
        {\bf Topics description} & {\bf Example tweet text} \\ 
        \midrule 
        
        \textbf{Affected Population} - Reports of injured, dead, missed, found, and the people affected due to the disaster event. & Latest: Mexico City Earthquake: At Least 225 Dead, Thousands Missing $\vert$ NBC Nightly News \\ \midrule
        
        \textbf{Infrastructure Damage} - Reports of any type of damage to infrastructures such as buildings, roads, bridges, power lines, communication poles, or vehicles. & RT @HumanityRoad: \#MexicoEarthquake - 300 houses damaged in \#Atzitzihuacan. \#hmrd \\ \midrule
        
        \textbf{Volunteering Support} - Reports of any type of rescue, volunteering, or donation efforts such as people receiving medical aid, donation of money, or services, etc. & RT @MLB: MLB to donate \$1 million to assist communities impacted by Hurricane Maria in PR and the earthquake in Mexico.\\ \midrule
        
        \textbf{Emergency Exercises} - Reports of any type of emergency preparedness drills and exercises for the disaster event & \#hagupit \#typhoon \#ruby coming to Philippines . Prepare storage ., drinks and feed . Stay safe. \\ \midrule
        
        \textbf{Early Warning} - Reports of any type of warning or alert signal issued related to the disaster event. & RT @NikaZaildar: NDMAs warning: There are chances of aftershocks in next 24 hours after todays \#earthquake \\ \midrule
        
        \textbf{Impact} - Reports of any type of aftermath activity (i.e., cleaning or rebuilding activities), population displacement, and disruption of economic activity. & Midwest ranchers face huge losses and massive cleanup after blizzards and flooding. @JournalStarNews \\ \midrule
        
        \textbf{Prayer} - Reports of any type of prayers, thoughts, and emotional support. & RT @crumpitout: Praying for all those affected by the earthquake in Mexico. Take care of each other. \\ \midrule
        
        \textbf{Supply Needs} - Reports of urgent needs or supplies such as food, water, clothing, money, medical supplies or blood. & Haiti needs money, food, medicine, construction materials and drinking water. \\ \midrule
        
        \textbf{Irrelevant} - The tweet does not fall into the given topics. & In Mexico, with a State that has failed in many areas, the people takes charge. This is huge! \#MexicoUnido \\ \bottomrule
    \end{tabular} }
\end{table*}

\subsection{Annotator Selection} \label{s:ansel}
\par  We observe that existing research works~\cite{rudra2015extracting, dutta2018ensemble, rudra2018extracting, nguyen2022towards} for ground-truth summary generation for disaster events do not provide any quality checking strategy for annotator selection. However, as the quality of ground-truth summary depends on the intuition and understanding of the annotators, we propose \textit{Quality Assessment Evaluation} to select annotators. For \textit{Quality Assessment Evaluation}, we evaluate annotators performance on a subset of tweets, $T^{'}$ from Hurricane Matthew~\footnote{https://en.wikipedia.org/wiki/Hurricane\_Matthew} ($D_2$) dataset. $T^{'}$ comprises of $2\%$ of tweets from each topic of a dataset. To handle fractions, we round-up $2\%$ of tweets. However, if the roundup results in zero tweet selection for a topic, we change it to $1$.   

\par In the \textit{Quality Assessment Evaluation}, we ask the annotators~\footnote{ The annotators are graduate students who belong to the age group of $20-30$, have good knowledge of English and are not a part of this project.} to \begin{enumerate*} [itemjoin = \quad] \item identify the topic given a tweet, and \item select the tweets from each topic into summary\end{enumerate*}. To identify the topic, we provide the annotators with a list of the possible topics along with descriptions and examples as shown in Table~\ref{table:Tsubthem}. On the basis of this provided information, the annotators assign the topic that seems the most relevant to the tweet text. To select the tweets into the summary, the annotator needs to identify the importance of a topic to determine its representation in summary and select the most representative tweets from each topic on the basis of the importance of that topic. We measure the annotator's performance on the basis of the generated summary quality through the objectives of text summarization~\cite{chakraborty2019tweet}, such as \textit{Coverage}, \textit{Relevance}, and \textit{Diversity}, through the opinion of a meta-annotator. \textit{Relevance} refers to the identification of the importance of each tweet with respect to a disaster event, \textit{Coverage} refers to the selection of the important aspects in summary, and \textit{Diversity} refers that all selected tweets in summary should have diverse/unique information, i.e., no two tweets convey the same information. We follow the existing summarization works~\cite{poddar2022caves, feigenblat2021tweetsumm}, where a meta-annotator scores the summary generated by an annotator in the range of $1$ (worst score) - $10$ (best score) on the basis of the fulfillment of the objectives, such as \textit{Coverage}, \textit{Relevance}, and \textit{Diversity}. A meta-annotator is a university graduate in the age group $20-30$, is well-versed in English and is conversant with Twitter. We consider an annotator to have passed the \textit{Quality Assessment Evaluation} if he/she scores more than $7$. For our ground-truth summary generation, we observed that $6$ out of $10$ annotators passed the \textit{Quality Assessment Evaluation}, we selected top-ranked $3$ annotators from them. We refer to these annotators as $P_1$, $P_2$, and $P_3$ in the rest of the paper.

\subsection{Summary Length} \label{s:sumlen}
We decide the length of the summary as $40$ on the basis of existing disaster summarization works~\cite{dutta2018ensemble, rudra2015extracting}. We do not follow any automated system to determine the number of tweets to be in summary on the basis of the disaster tweets. 

\section{Results and Discussions}\label{s:res}
\par In this Section, we evaluate the effectiveness of PORTRAIT by comparing the ground-truth summary generated by PORTRAIT with the ground-truth summary generated by an existing semi-automated approach~\cite{nguyen2018tsix} and the existing research works specific to disaster events~\cite{rudra2015extracting, dutta2018ensemble, rudra2018identifying}. We refer to the summary generated by existing semi-automated approach as \textit{Semi-automated Summary}, existing approaches specific to disaster events as \textit{Baseline Summary} and the summary generated by PORTRAIT as \textit{Proposed Summary}. As \textit{Semi-automated Summary} and \textit{Baseline Summary} require at least $3$ annotators, we employ $3$ annotators for both of them. We refer the annotators for \textit{Semi-automated Summary} as $S_1$, $S_2$ and $S_3$ and  for the \textit{Baseline Summary} as $B_1$, $B_2$ and $B_3$. As previously discussed, we refer to the annotators for PORTRAIT as $P_1$, $P_2$ and $P_3$. 

\par 
We have considered $3$ metrics, namely \textit{Coverage}, \textit{Relevance}, and \textit{Diversity} for performance evaluation of PORTRAIT. For qualitative comparison, we employ $3$ meta-annotators for the subjective understanding of each summary on the basis of considered metrics \textit{Coverage}, \textit{Relevance}, and \textit{Diversity} in subsection~\ref{s:analy}. Additionally, we compare the summaries through the quantitative understanding of  \textit{Coverage}, \textit{Relevance}, and \textit{Diversity} in subsection~\ref{s:justifi}. 
We, finally, provide a case study where we evaluate the existing summarization approaches on the ground-truth summaries generated by PORTRAIT for $D_1-D_5$ datasets in subsection~\ref{s:exp}. 

\begin{landscape}
\begin{table}
    \caption{We show the aggregate (average) coverage score, relevance score, and diversity score of the \textit{Proposed Summary}, \textit{Semi-automated Summary}, and \textit{Baseline Summary} of all the $3$ annotators for $D_1-D_5$ datasets.}
    \label{table:eval}
    \smallskip
    \resizebox{\linewidth}{!}{\begin{tabular} {cccccccccc}
        \toprule
        {\bf Dataset} & \multicolumn{3}{c}{\bf Proposed Summary} & \multicolumn{3}{c}{\bf Semi-automated Summary} & \multicolumn{3}{c}{\bf Baseline Summary} \\ \cline{2-10}
        & {\bf Aggregate} & {\bf Aggregate} & {\bf Aggregate} & {\bf Aggregate} & {\bf Aggregate} & {\bf Aggregate} & {\bf Aggregate} & {\bf Aggregate} & {\bf Aggregate}\\
        & {\bf coverage} & {\bf relevance} & {\bf diversity} & {\bf coverage} & {\bf relevance} & {\bf diversity} & {\bf coverage} & {\bf relevance} & {\bf diversity}\\
        & {\bf score} & {\bf score} & {\bf score} & {\bf score} & {\bf score} & {\bf score} & {\bf score} & {\bf score} & {\bf score}\\ \midrule 
        
        $D_1$   & 4.70 & 4.84 & 4.80 & 4.22 & 4.36 & 3.94 & 3.94 & 4.19 & 3.86 \\ 
        $D_2$   & 4.49 & 4.75 & 4.71 & 3.67 & 4.44 & 3.67 & 3.78 & 4.22 & 3.38 \\ 
        $D_3$   & 4.58 & 4.69 & 4.89 & 4.47 & 3.64 & 3.83 & 4.33 & 3.31 & 3.94 \\ 
        $D_4$   & 4.83 & 4.25 & 4.81 & 4.17 & 4.06 & 3.47 & 4.05 & 3.47 & 3.75 \\ 
        $D_5$   & 4.64 & 4.78 & 4.85 & 4.00 & 3.67 & 4.25 & 3.69 & 3.44 & 4.42 \\ \bottomrule
    \end{tabular}}
\end{table}

\begin{table}
    \caption{We show the number of topics in a dataset, \textit{Proposed Summary}, \textit{Semi-automated Summary}, and \textit{Baseline Summary} of all $3$ annotators, and the topics missing in both the summaries for $D_1-D_5$ datasets. (Note: \# represents number in this table)}
    \label{table:catcoverage}
    \smallskip
    \resizebox{\linewidth}{!}{\begin{tabular} {cccccccccccccc}
        \toprule
        {\bf Dataset} & \bf{\# of topics}  & \multicolumn{4}{c}{\bf Proposed Summary} & \multicolumn{4}{c}{\bf Semi-automated Summary} & \multicolumn{4}{c}{\bf Baseline Summary}\\ \cline{3-14}
                      
                      & \bf{in a}   & \multicolumn{3}{c}{\bf \# of topics}  & \bf{Missed topics} & \multicolumn{3}{c}{\bf \# of topics} & \bf{Missed topics} & \multicolumn{3}{c}{\bf \# of topics} & \bf{Missed topics} \\\cline{3-5} \cline{7-9} \cline{11-13}
                      
                      &  \bf{dataset} &  {\bf $P_1$} & {\bf $P_2$} & {\bf $P_3$} & \bf{across annotators} &  {\bf $S_1$} & {\bf $S_2$} & {\bf $S_3$} & \bf{across annotators} &  {\bf $B_1$} & {\bf $B_2$} & {\bf $B_3$} & \bf{across annotators} \\\midrule 
        
        $D_1$   & 9  & 8  & 8  & 8  & Infrastructure Damage & 7 & 6 & 7 & Infrastructure Damage & 5 & 5 & 7 & Emotional Distress, Infrastructure Damage \\ 
        $D_2$   & 9  & 8  & 7  & 7  & Emotional Distress    & 6 & 5 & 7 & Humanitarian Event, Emotional Distress & 6 & 6 & 7 & Emotional Distress \\ 
        $D_3$   & 9  & 8  & 7  & 8  &        -              & 5 & 5 & 6 & Emergency Exercise, Humanitarian Event, Emotional Distress & 4 & 5 & 5 & Early Warning, Emotional Distress, Humanitarian Event \\ 
        $D_4$   & 10 & 10 & 10 & 10 &        -              & 6 & 7 & 7 & Humanitarian Event, Emotional Distress & 6 & 8 & 8 & Emotional Distress \\ 
        $D_5$   & 8  & 7  & 7  & 7  & Humanitarian Event    & 3 & 5 & 3 & Emergency Exercise, Humanitarian Event, Early Warning, Emotional Distress & 5 & 7 & 6 & Emergency Exercise, Humanitarian Event \\ \bottomrule
    \end{tabular}}
\end{table}
\end{landscape}

\subsection{Qualitative Comparison} \label{s:analy}
\par Qualitative assessment is a well-accepted method to evaluate summary quality. For quality assessment, we gave the input tweets related to the disaster event, \textit{Proposed Summary}, \textit{Semi-automated Summary} and \textit{Baseline Summary} to $3$ meta-annotators. We asked the meta-annotator to rate the summary on the basis of three factors, namely  \textit{Coverage}, \textit{Relevance}, and \textit{Diversity}. We also provide annotators with the definition of these three factors as follows - \begin{enumerate*} [itemjoin = \quad] \item \textit{Coverage} indicates the percentage of important sub-events/aspects present in the input tweets that are covered in summary, \item \textit{Relevance} of a tweet indicates how much relevant a tweet is with respect to the corresponding disaster event. So, the \textit{Relevance} of a summary depends on the percentage of tweets in the summary which are relevant to the disaster event, and \item \textit{Diversity} indicates that tweets in summary comprise of diverse information \end{enumerate*}. We asked the meta-annotators to rate the \textit{Proposed Summary}, \textit{Semi-automated Summary} and \textit{Baseline Summary} on each factor in the range of $1$ (worst rating) - $5$ (best rating) for the $5$ disaster datasets. We also asked them to choose a fractional score if required. In Table~\ref{table:eval}, we show the aggregated (average) score of $3$ annotators for all the three factors on $5$ datasets. We observe that the aggregated score for all factors are more than $4$ for all the $5$ datasets for the \textit{Proposed Summary}. Additionally, we observe that the Aggregated coverage score for all datasets ranges between $4.49$-$4.83$, the relevance score between $4.25$-$4.84$ and the diversity score between $4.71$-$4.85$ for the \textit{Proposed Summary} whereas the Aggregated coverage score ranges between $3.67$-$4.47$ and $3.69$-$4.33$, the relevance score between $3.64$-$4.44$ and $3.31$-$4.22$, the diversity score between $3.47$-$4.25$ and $3.38$-$4.42$ for \textit{Semi-automated Summary} and \textit{Baseline Summary} respectively. Therefore, our observations indicate that the quality of the \textit{Proposed Summary} is very high.

\subsection{Quantitative Comparison} \label{s:justifi}
In this Section, we present the quantitative comparison among \textit{Proposed Summary}, \textit{Semi-automated Summary} and \textit{Baseline Summary} in terms of coverage, relevance and diversity.

\subsubsection*{\textbf{Coverage}} As mentioned earlier that a good quality summary should cover all the important sub-events/aspects/topics of the event.
In order to understand this, we compare the topic coverage among all the summaries. We utilize the topics identified by PORTRAIT in Section~\ref{s:ch} for the \textit{Proposed Summary}, \textit{Semi-automated Summary} and \textit{Baseline Summary}. We show the number of topics for $D_1-D_5$ datasets in Table~\ref{table:catcoverage}. We found that there is atmost one topic is not captured in \textit{Proposed Summary} with respect to all the topics in input tweets. However, on observing the tweets related to the topic which is not captured, we found that both the number of tweets and relevance of those tweets with respect to the disaster is very low. For example, for $D_1$ which comprises of tweets related to the disaster event, \textit{Los Angeles International Airport Shooting}~\footnote{https://en.wikipedia.org/wiki/2013\_Los\_Angeles\_International\_Airport\\\_shooting}, we found that there is no tweet which belongs to the topic, \textit{Infrastructure Damage} in \textit{Proposed Summary}. However, as the event name suggests, there was no major infrastructure damage during \textit{Los Angeles International Airport Shooting}, and the number of tweets that belongs to this topic was very low, i.e., $1$ tweet. Additionally, we observe that there was no tweet that belonged to \textit{Infrastructure Damage} in the  \textit{Semi-automated Summary} and \textit{Baseline Summary} for $D_1$. However, there were other topics, such as, \textit{Emotional Distress} which comprised of $1$ tweet and $5$ tweets for \textit{Hurricane Matthew}~\footnote{https://en.wikipedia.org/wiki/Hurricane\_Matthew} ($D_2$) and \textit{Puebla Mexico Earthquake}~\footnote{https://en.wikipedia.org/wiki/2017\_Puebla\_earthquake} ($D_3$), respectively, \textit{Humanitarian Event} which comprised of $5$ tweets for \textit{Midwestern U.S. Floods}~\footnote{https://en.wikipedia.org/wiki/2019\_Midwestern\_U.S.\_floods} ($D_5$), etc., were missing in both the \textit{Semi-automated Summary} and \textit{Baseline Summary}. We observe similar findings across all the $5$ datasets that the topic which was not captured by \textit{Proposed Summary} was not captured by either \textit{Semi-automated Summary} or \textit{Baseline Summary}. However, both the \textit{Semi-automated Summary} and \textit{Baseline Summary} did not capture several additional topics which were covered by \textit{Proposed Summary}. Therefore, our observations show  that \textit{Proposed Summary} has a higher topic coverage than both \textit{Semi-automated Summary} and \textit{Baseline Summary} across all datasets.

\subsubsection*{\textbf{Relevance}}\label{s:relav} Summary should ensure that the relevant tweets of the disaster event are captured. In order to understand the relevance of each tweet, we ask meta-annotators to annotate all the tweets in the input dataset with \textit{relevance label}, which are \textit{high}, \textit{medium} or \textit{low} on the basis of their wisdom and intuition. Additionally, we ask the meta-annotator to provide explainables or explanations behind their decision of the \textit{relevance label} for each tweet to support \textit{relevance label} annotation. A meta-annotator has good knowledge of English and was not a part of this project. We show a few examples of this annotation in Table~\ref{table:relavexamp}. In order to evaluate \textit{Proposed Summary} with the \textit{Semi-automated Summary} and \textit{Baseline Summary} with respect to \textit{relevance}, we check the distribution of \textit{high}, \textit{medium} and \textit{low} \textit{relevance label} tweets in the respective summaries. We show the percentage of each \textit{relevance label} for all the summaries of all $3$ annotators for $D_1-D_5$ datasets in Table~\ref{table:informlabel}. Our observation indicates that $82.50\%-92.50\%$ of tweets in the \textit{Proposed Summary} have \textit{high} \textit{relevance labels}, whereas $22.50\%-75.00\%$ of tweets in the \textit{Semi-automated Summary} and $30.00\%-70.00\%$ of tweets in the \textit{Baseline Summary} have \textit{high} \textit{relevance labels}. Similarly, $7.50\%-17.50\%$ of tweets in the \textit{Proposed Summary} have \textit{medium} \textit{relevance labels}, whereas $2.50\%-30.00\%$ of tweets in the \textit{Semi-automated Summary} and $7.50\%-22.50\%$ of tweets in the \textit{Baseline Summary} have \textit{medium} \textit{relevance labels}. We further observe that none of the tweets in the \textit{Proposed Summary} has \textit{low} \textit{relevance labels} across the disasters, whereas $15.00\%-62.50\%$ of tweets in the \textit{Semi-automated Summary} and $22.50\%-65.00\%$ of the tweets in the \textit{Baseline Summary} have \textit{low} \textit{relevance labels}. Therefore, based on this observation, we can say that PORTRAIT ensures more \textit{high} \textit{relevance labels} tweets and no \textit{low} \textit{relevance labels} tweets in summary.

\begin{table*}[!ht]
    \caption{We show a few examples of tweets and corresponding \textit{relevance label} annotations with explanations.}
    \label{table:relavexamp}
    \resizebox{\textwidth}{!}{\begin{tabular} {p{0.45\linewidth}p{0.35\linewidth}c}
        \toprule
        {\bf Tweet text} & {\bf Explanation}& {\bf Relevance label} \\ 
        \midrule 
        
        \#Jamaica Haiti: Hurricane Matthew: 350,000 people in need of assistance, 15,623 are displaced  \#crisismanagement	&   350,000 people need assistance 15,623 displaced  &  High \\ \midrule
        
        @christian\_aid staff homes badly damaged \#HurricaneMatthew At least six feared dead in Haiti as violent storm hits	&   At least six dead   & 	High \\ \midrule

        A Hurricane Warning is issued for Jamaica \&amp; much of Haiti. A \#Hurricane Watch is now in effect for SE Cuba.  \#Matthew	&   Hurricane Warning issued for Jamaica	&   High \\ \midrule

        T-Mobile offering free calling and texting to countries affected by Hurricane Matthew  via @tmonews @HavServe \#Haiti   &   offering free calling texting to affected by Hurricane Matthew  &  Medium \\ \midrule

        RT @KSNTNews: Haiti is starting to assess damage from Hurricane Matthew &	Haiti assess damage from Hurricane Matthew  &	Medium  \\ \midrule

        If you really want to know what Clintons did/didnt do in \#Haiti \&amp; how US aid works @KatzOnEarth cuts thru the b.s. \#Matthew   &	want to know what Clintons did/didnt   & 	Low \\ \midrule

        Haiti Floods and Flooding: Hurricane Matthew $\vert$ ImperialHipHop   &	Haiti Floods and Flooding   & 	Low \\ \bottomrule
    \end{tabular}}
\end{table*}

\begin{table*}[!ht]
    \caption{We show the percentage number of tweets of each \textit{relevance label}, such as \textit{high}, \textit{medium}, and \textit{low} for \textit{Proposed Summary}, \textit{Semi-automated Summary}, and \textit{Baseline Summary} of all the $3$ annotators for $D_1-D_5$ datasets.}
    \label{table:informlabel}
    \centering
    \begin{tabular}{cccccccccc}\toprule
        \textbf{Dataset} & \multicolumn{9}{c}{\textbf{Proposed Summary}} \\ \cline{2-10}
        & \multicolumn{3}{c}{$\bf P_1$} & \multicolumn{3}{c}{$\bf P_2$} & \multicolumn{3}{c}{$\bf P_3$} \\ \cline{2-10}
        & \textbf{High} & \textbf{Medium} & \textbf{Low} & \textbf{High} & \textbf{Medium} & \textbf{Low} & \textbf{High} & \textbf{Medium} & \textbf{Low} \\ \midrule 
        $D_1$   & 85.00\% & 15.00\% & - & 82.50\% & 17.50\% & - & 82.50\% & 17.50\% & - \\ 
        $D_2$   & 92.50\% & 7.50\%  & - & 90.00\% & 10.00\% & - & 85.00\% & 15.00\% & - \\ 
        $D_3$   & 90.00\% & 10.00\% & - & 92.50\% & 7.50\%  & - & 87.50\% & 12.50\% & - \\ 
        $D_4$   & 87.50\% & 12.50\% & - & 87.50\% & 12.50\% & - & 82.50\% & 17.50\% & - \\ 
        $D_5$   & 87.50\% & 12.50\% & - & 87.50\% & 12.50\% & - & 87.50\% & 12.50\% & - \\ \bottomrule
    \end{tabular}
    \bigskip
    
    \begin{tabular}{cccccccccc} \toprule
        \textbf{Dataset} & \multicolumn{9}{c}{\textbf{Semi-automated Summary}} \\ \cline{2-10}
        & \multicolumn{3}{c}{$\bf S_1$} & \multicolumn{3}{c}{$\bf S_2$} & \multicolumn{3}{c}{$\bf S_3$} \\ \cline{2-10}
        & \textbf{High} & \textbf{Medium} & \textbf{Low} & \textbf{High} & \textbf{Medium} & \textbf{Low} & \textbf{High} & \textbf{Medium} & \textbf{Low} \\ \midrule 
        $D_1$   & 42.50\% &  7.50\% & 50.00\% & 30.00\% & 15.00\% & 55.00\% & 22.50\% & 15.00\% & 62.50\% \\ 
        $D_2$   & 70.00\% & 10.00\% & 20.00\% & 65.00\% & 07.50\% & 27.50\% & 67.50\% &  7.50\% & 25.00\% \\ 
        $D_3$   & 60.00\% & 12.50\% & 44.00\% & 70.00\% & 15.00\% & 15.00\% & 75.00\% & 2.50\%  & 22.50\% \\ 
        $D_4$   & 67.50\% & 10.00\% & 22.50\% & 57.50\% & 12.50\% & 30.00\% & 57.50\% & 12.50\% & 30.00\% \\ 
        $D_5$   & 55.00\% &  7.50\% & 32.50\% & 45.00\% & 30.00\% & 25.00\% & 37.50\% & 17.50\% & 45.00\% \\ \bottomrule
    \end{tabular}
    \bigskip
    
    \begin{tabular}{cccccccccc}\toprule
        \textbf{Dataset} & \multicolumn{9}{c}{\textbf{Baseline Summary}} \\ \cline{2-10}
        & \multicolumn{3}{c}{$\bf B_1$} & \multicolumn{3}{c}{$\bf B_2$} & \multicolumn{3}{c}{$\bf B_3$} \\ \cline{2-10}
        & \textbf{High} & \textbf{Medium} & \textbf{Low} & \textbf{High} & \textbf{Medium} & \textbf{Low} & \textbf{High} & \textbf{Medium} & \textbf{Low} \\ \midrule 
        $D_1$   & 35.00\% & 12.50\% & 52.50\% & 40.00\% & 12.50\% & 47.50\% & 30.00\% & 5.00\%  & 65.00\% \\ 
        $D_2$   & 50.00\% & 15.00\% & 35.00\% & 42.50\% & 10.00\% & 47.50\% & 47.50\% & 20.00\% & 32.50\% \\ 
        $D_3$   & 50.00\% & 15.00\% & 35.00\% & 62.50\% & 15.00\% & 22.50\% & 70.00\% & 7.50\%  & 22.50\% \\ 
        $D_4$   & 50.00\% & 12.50\% & 37.50\% & 57.50\% & 10.00\% & 32.50\% & 47.50\% & 7.50\%  & 45.00\% \\ 
        $D_5$   & 35.00\% & 22.50\% & 42.50\% & 50.00\% & 10.00\% & 40.00\% & 45.00\% & 17.50\% & 37.50\% \\ \bottomrule
    \end{tabular}
\end{table*}

\subsubsection*{\textbf{Diversity}} A summary should ensure that the tweets selected in the summary capture diverse information. In order to calculate the diversity of the summary, $S$, we calculate the aggregate (average) diversity score, which is the average of diversity between each pair of tweets, say $T_i$ and $T_j$, $Div(T_i,T_j)$ as $AvgDiv(S)$. We calculate $Div(T_i,T_j)$ as :

\begin{align}
   Div(T_i,T_j) = 1 - Sim(T^{x}_i,T^{x}_j)
    \label{eq:divS}
\end{align}

where, $Sim(T^{x}_i,T^{x}_j)$ represents the semantic similarity between a pair of tweets explainables, $T^{x}_i$ and $T^{x}_j$ of $T_i$ and $T_j$, respectively, by:

\begin{align}
  Sim(T^{x}_i,T^{x}_j) = \frac{\vec{E_i} \cdot \vec{E_j}}{\vert \vec{E_i} \vert \ \vert \vec{E_j} \vert}
    \label{eq:SemSim}
\end{align}

where, $\vec{E_i}$ and $\vec{E_j}$ are the embedding of $T^{x}_i$ and $T^{x}_j$ respectively. We calculate $\vec{E_i}$ and $\vec{E_j}$ as the average of the values of the tweet \textit{explainable} keywords embedding of $T^{x}_i$ and $T^{x}_j$, respectively. We consider the embedding of an \textit{explainable} keyword of a tweet using a pre-train Word2Vec model provided by CrisisNLP~\cite{imran2016twitter}, which is trained on $52$ million crisis-related messages of various disaster events. However, as tweets do not inherently contain \textit{explainables} which can represent the information present in the tweet about a disaster event, we rely on the \textit{explainables} provided by meta-annotators (as discussed in subsection~\ref{s:relav}) of all the tweets in summary. We calculate $AvgDiv(S)$ of the \textit{Proposed Summary}, \textit{Semi-automated Summary} and \textit{Baseline Summary} of $3$ meta-annotators for all the $5$ datasets. Our observations as shown in Table~\ref{table:diveval} indicate that $AvgDiv(S)$ ranges from $0.45-0.69$ in \textit{Proposed Summary}, whereas it ranges from $0.40-0.66$ in \textit{Semi-automated Summary} and $0.43-0.66$ in \textit{Baseline Summary}. Therefore, \textit{Proposed Summary} obtains $2.62\%-8.12\%$ and $2.28\%-5.68\%$ higher aggregate diversity score as compared to \textit{Semi-automated Summary} and \textit{Baseline Summary}, respectively, which implies PORTRAIT ensures more diverse tweets in summary than existing ground-truth summary techniques.

\begin{table}[ht!]
    \caption{We show the aggregate (average) diversity score of the \textit{Proposed Summary}, \textit{Semi-automated Summary}, and \textit{Baseline Summary} of all the $3$ annotators for $D_1-D_5$ datasets.}
    \label{table:diveval}
    \centering
    \resizebox{\textwidth}{!}{\begin{tabular} {cccc}
        \toprule
        {\bf Dataset} & \bf Proposed Summary & \bf Semi-automated Summary & \bf Baseline Summary \\ \cline{2-4}
            & {\bf Aggregate diversity} & {\bf Aggregate diversity} & {\bf Aggregate diversity} \\
            & {\bf score} & {\bf score} & {\bf score} \\ \midrule 
        
        $D_1$   & 0.5711 & 0.5565 & 0.5404 \\ 
        $D_2$   & 0.6918 & 0.6611 & 0.6595 \\ 
        $D_3$   & 0.5382 & 0.5093 & 0.5175 \\ 
        $D_4$   & 0.5325 & 0.5122 & 0.5206 \\ 
        $D_5$   & 0.4543 & 0.4202 & 0.4342 \\ \bottomrule
    \end{tabular}}
\end{table}

\subsection{Case Study : Evaluation of Existing Summarization Approaches } \label{s:exp}
\par In this subsection, we initially discuss the details of the existing state-of-the-art summarization approaches. Then, we provide a performance comparison of these approaches on the ground-truth summaries generated by PORTRAIT for $5$ disaster datasets.

\subsubsection{Existing Summarization Approaches}\label{s:baseline}
\par We segregate these approaches into \textit{content-based}, \textit{graph-based}, \textit{matrix factorization-based}, \textit{semantic similarity-based }, \textit{ontology-based} and \textit{deep learning-based} approaches. We select few prominent tweet summarization approaches from each type which we discuss next.

\begin{enumerate}
    
    \item \textit{Content-based Approaches:} We discuss the existing content-based summarization approaches as follows:
        \begin{enumerate}
            \item \textit{LUHN}: Luhn et al.~\cite{luhn1958automatic} propose a frequency-based summarization approach which initially determines the term frequency score of each word in a document (after removing stopwords and stemming) and then, generates a summary by the selection of those sentences into summary which has the highest frequency scoring words. 
            
            \item \textit{SumBasic}: Nenkova et al.~\cite{nenkova2005impact} initially identify the probability of occurrence of each word in a document and then, select those tweets into summary which has the words with the maximum probability of occurrence.
            
            \item \textit{COWTS}: Rudra et al.~\cite{rudra2015extracting} initially calculate the score of each keyword (i.e., noun, main verb and numerals) using TF-IDF and then, select a tweet into summary if it contains the keywords with maximum score. 
    
            \item \textit{DEPSUB}: Rudra et al.~\cite{rudra2018identifying} initially identify the sub-events from the tweets and select those representative tweets from each sub-event into summary, which can ensure maximum coverage of the sub-event. 
            
        \end{enumerate}

    \item \textit{Graph-based Approaches:} We discuss the existing graph-based summarization approaches as follows:
        \begin{enumerate}
            \item \textit{Cluster Rank}: Garg et al.~\cite{garg2009clusterrank} initially segments a document into clusters followed by PageRank~\cite{page1999pagerank}  algorithm to identify the tweets from each cluster to be selected into summary.  
            
            \item \textit{LexRank}: Erkan et al.~\cite{erkan2004lexrank} propose initially constructs a graph where the nodes are the sentences and the edges represent the cosine similarity between each pair of sentences and finally, selects those sentences which have the highest Eigenvector~\cite{borgatti2005centrality} centrality score into the summary. 
            
            \item \textit{$EnSum$}: Dutta et al.~\cite{dutta2018ensemble} propose an ensemble graph-based tweet summarization approach, \textit{$EnSum$} in which they initially identify the tweets by $9$ summarization algorithms and then, create a tweet graph that comprises of these tweets as nodes and edges represent their similarity. Finally, they select tweets with the highest representativeness score from the tweet graph in summary. 
            
            \item \textit{COWEXABS}: Rudra et al.~\cite{rudra2019summarizing} propose initially identify the most relevant disaster-specific keywords and then, select those tweets into the summary that provide maximum information coverage of these keywords.
                       
            \item \textit{MEAD}: Radev et al.~\cite{radev2004mead} propose a centroid-based summarization approach which initially identifies the clusters by agglomerative clustering and then, selects tweets from each cluster into the summary on the basis of  centrality score and diversity score. 
        \end{enumerate}

    \item \textit{Matrix factorization-based Approaches:} We discuss the most popular matrix factorization-based summarization approaches. 
    
    \begin{enumerate}
    
        \item \textit{LSA}: Gong et al.~\cite{gong2001generic} propose a document summarization approach, \textit{LSA}, which selects the tweets with the largest eigenvalues after Singular Value Decomposition (SVD) of the keyword matrix created from all the tweets. 
        
        \item \textit{SumDSDR}: He et al.~\cite{he2012document} propose a data reconstruction-based document summarization approach. \textit{SumDSDR} measure the relationship among the sentences using linear reconstruction and non-linear reconstruction objective functions and then create a summary by minimizing the reconstruction error. 
    \end{enumerate}
    
    \item \textit{Ontology-based Approach:}  Garg et al.~\cite{garg2022ontodsumm} propose an ontology-based tweet summarization approach, \textit{OnntoDSumm}, which initially identifies the category of each tweet using an ontology-based pseudo-relevance feedback approach followed by determination of the importance of each category with respect to a disaster. Finally, select the representative tweets from each category based on the disaster-specific maximal marginal relevance (DMMR) based approach to create a summary.
    
    \item \textit{Deep learning-based Approach:} Nguyen et al.~\cite{nguyen2022towards} propose disaster-specific abstractive tweet summarization approach, \textit{RATSUM}, which identify the key-phrases present in tweets using a pre-trained BERT model~\cite{liu2019text} and then generate the word summary by maximizing the coverage of key-phrases in the final summary. For our experiments, we select those tweets into the summary, which provides the maximum coverage of key phrases in the final summary.
     
\end{enumerate}

\subsubsection{Comparison Results and Discussions}\label{s:result}
\par To evaluate the performance of the various state-of-the-art summarization approaches, we compare the summary generated by different approaches using ROUGE-N~\cite{lin2004rouge} scores. ROUGE-N score is a well-known measure in text summarization tasks, which computes the score on the basis of overlapping words between the system-generated summary and the ground-truth summary. We use F1-score for $3$ different variants of the ROUGE-N score, i.e., N=$1$, $2$ and L, respectively. The higher the ROUGE score, better is the quality of the summary. Our observations from Table~\ref{table:results1} indicate that \textit{OntoDSumm} ensures the best ROUGE-N F1-scores on $D_1-D_5$ followed by \textit{RATSUM}. The reason behind the high performance is that \textit{OntoDSumm} utilizes ontology knowledge with respect to each topic to identify the importance of each tweet in a topic. Additionally, it captures the representation of each topic in summary and handles the information diversity in summary tweets. Further, our observation indicates that \textit{RATSUM} ensures the best ROUGE-N F1-scores on $D_1$ and $D_3$ followed by \textit{LexRank}. The reason for the high performance is that \textit{RATSUM} better captures the content and context information presents in the tweet to predict the tweet importance. However, it does not cover the information diversity in summary tweets. The performance of \textit{MEAD} and \textit{COWTS} are the worst for $D_1$ and $D_3-D_5$, and $D_2$, respectively, because they did not cover category representation and information diversity in summary.

\begin{table*}
    \caption{F1-score of ROUGE-1, ROUGE-2 and ROUGE-L score of the summaries generated by various state-of-the-art summarization approaches on $D_1$-$D_5$ datasets.}
    \label{table:results1}
    \centering 
    \resizebox{\textwidth}{!}{\begin{tabular}{cccccccccc}
        \toprule
        
        \textbf{Approach} & \multicolumn{3}{c}{\textbf{$D_1$}} & \multicolumn{3}{c}{\textbf{$D_2$}} & \multicolumn{3}{c}{\textbf{$D_3$}}  \\ \cline{2-10}
        
        & \textbf{ROUGE-1} & \textbf{ROUGE-2} & \textbf{ROUGE-L} & \textbf{ROUGE-1} & \textbf{ROUGE-2} & \textbf{ROUGE-L} & \textbf{ROUGE-1} & \textbf{ROUGE-2} & \textbf{ROUGE-L} \\ \midrule
        
        $Cluster Rank$  & 0.46 & 0.21 & 0.31 &      0.51 & 0.17 & 0.27   & 0.58 & 0.25 & 0.29 \\ 
        $Lex Rank$      & 0.54 & 0.26 & 0.32 &      0.54 & 0.20 & 0.29   & 0.58 & 0.24 & 0.30 \\ 
        $LSA$           & 0.49 & 0.15 & 0.24 &      0.49 & 0.16 & 0.24   & 0.48 & 0.14 & 0.22 \\ 
        $LUHN$          & 0.52 & 0.18 & 0.26 &      0.50 & 0.15 & 0.24   & 0.58 & 0.19 & 0.27 \\ 
        $MEAD$          & 0.38 & 0.10 & 0.22 &      0.49 & 0.12 & 0.23   & 0.47 & 0.14 & 0.24 \\ 
        $SumBasic$      & 0.55 & 0.20 & 0.28 &      0.54 & 0.17 & 0.25   & 0.56 & 0.24 & 0.28 \\ 
        $SumDSDR$       & 0.55 & 0.26 & 0.33 &      0.54 & 0.20 & 0.28   & 0.57 & 0.23 & 0.29 \\ 
        $COWTS$         & 0.51 & 0.22 & 0.27 &      0.47 & 0.12 & 0.22   & 0.50 & 0.18 & 0.23 \\ 
        $COWEXABS$      & 0.51 & 0.23 & 0.30 &      0.54 & 0.22 & 0.26   & 0.55 & 0.23 & 0.28 \\ 
        $DEPSUB$        & 0.51 & 0.20 & 0.28 &      0.52 & 0.15 & 0.23   & 0.55 & 0.20 & 0.27 \\ 
        $EnSum$         & 0.47 & 0.16 & 0.26 &      0.50 & 0.15 & 0.25   & 0.54 & 0.21 & 0.26 \\ 
        $OntoDSumm$     & {\bf0.60} & {\bf0.29} & {\bf0.37} &      {\bf0.57} & {\bf0.22} & {\bf0.30}   & {\bf0.61} & {\bf0.27} & {\bf0.31} \\ 
        $RATSUM$        & 0.58 & 0.27 & 0.35 &      0.55 & 0.20 & 0.28   & 0.60 & 0.26 & 0.30 \\ \bottomrule
    \end{tabular}}

    \bigskip
    \centering 
    \begin{tabular}{ccccccc}
        \toprule
        \textbf{Approach} & \multicolumn{3}{c}{\textbf{$D_4$}} & \multicolumn{3}{c}{\textbf{$D_5$}} \\ \cline{2-7}
        & \textbf{ROUGE-1} & \textbf{ROUGE-2} & \textbf{ROUGE-L} & \textbf{ROUGE-1} & \textbf{ROUGE-2} & \textbf{ROUGE-L} \\ \midrule
        
        $Cluster Rank$  & 0.44 & 0.12 & 0.22    & 0.50 & 0.17 & 0.24 \\ 
        $Lex Rank$      & 0.38 & 0.11 & 0.22    & 0.51 & 0.17 & 0.24 \\ 
        $LSA$           & 0.50 & 0.12 & 0.20    & 0.44 & 0.09 & 0.19 \\ 
        $LUHN$          & 0.51 & 0.13 & 0.22    & 0.51 & 0.15 & 0.21 \\ 
        $MEAD$          & 0.42 & 0.06 & 0.18    & 0.49 & 0.10 & 0.20 \\ 
        $SumBasic$      & 0.48 & 0.12 & 0.22    & 0.52 & 0.15 & 0.21 \\ 
        $SumDSDR$       & 0.42 & 0.15 & 0.24    & 0.45 & 0.12 & 0.21 \\ 
        $COWTS$         & 0.44 & 0.08 & 0.18    & 0.46 & 0.14 & 0.24 \\ 
        $COWEXABS$      & 0.46 & 0.13 & 0.23    & 0.22 & 0.05 & 0.20 \\ 
        $DEPSUB$        & 0.50 & 0.14 & 0.24    & 0.51 & 0.15 & 0.22 \\ 
        $EnSum$         & 0.48 & 0.12 & 0.21    & 0.49 & 0.13 & 0.21 \\ 
        $OntoDSumm$     & {\bf0.53} & {\bf0.17} & {\bf0.26}    & {\bf0.56} & {\bf0.19} & {\bf0.27} \\ 
        $RATSUM$        & 0.46 & 0.13 & 0.22    & 0.55 & 0.15 & 0.22 \\ \bottomrule
    \end{tabular}
\end{table*}

\section{Conclusions and Future works}\label{s:con}

\par In this paper, we propose a hybrid approach, PORTRAIT, which partially automates the extractive ground-truth summary generation for disaster events. Therefore, by this hybrid approach, we can handle both of the inherent challenges for ground-truth summary generation, i.e., reduce the effort and time of human annotators and ensure consistency in summary irrespective of the annotators. In order to understand whether the adoption of automation and reduction of human effort and time in ground-truth summary generation affects the ground-truth summary quality, we compare the performance of PORTRAIT with the existing approaches for ground-truth summary generation by $3$ annotators both quantitatively and qualitatively on $5$ disaster events datasets. Our observations indicate that the summary quality by PORTRAIT is better than the existing approaches by both quantitative and qualitative measures. Additionally, we observed that the variance among the ground-truth summaries generated by the $3$ annotators for $5$ disaster events datasets is very less, which indicates that PORTRAIT can ensure consistent summaries across annotators. Further, on the basis of these observations, we can explore a new direction in the ground-truth summary generation for disaster events such that there is no requirement for multiple annotators.

\par  Apart from PORTRAIT, in this paper, we generate and publically provide ground-truth summaries for $5$ different disaster datasets of different types, including earthquake, hurricane, flood, and mass shootings, which occurred in various countries, such as the United States of America, Haiti, Mexico, and Pakistan. We believe this will help in the development and evaluation of disaster tweet summarization approaches. Additionally, we perform a case study where we study and evaluate the performance of $13$ state-of-the-art summarization approaches on these $5$ disaster datasets summaries using ROUGE-N F1-scores.

\section*{Acknowledgements}
\par 
The authors would like to express their gratitude to the annotators who provided us with the ground-truth summary. The authors thank Aditya Kumar, Juhi Rani, and Thiyagura Pragathi for their help in the implementation of some existing summarization approaches. 

 

\bibliographystyle{IEEEtran}
\bibliography{journal_bib.bib}
\begin{IEEEbiography}[{\includegraphics[width=1in,height=1.25in,clip,keepaspectratio]{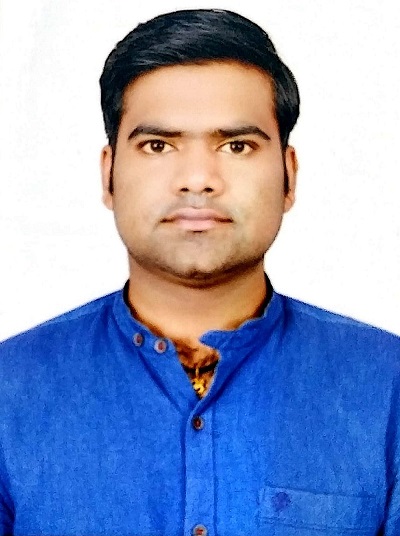}}]{Piyush Kumar Garg} is a PhD Scholar with the Department of Computer Science and Engineering, Indian Institute of Technology, Patna, India. He received the M.Tech. degree from IIT(ISM) Dhanbad, India in 2018 and B.Tech degree from the College of Technology and Engineering, Udaipur, India in 2015. His current research interests include social network analysis, crisis response, and information retrieval.
\end{IEEEbiography}
\begin{IEEEbiography}[{\includegraphics[width=1in,height=3in,clip,keepaspectratio]{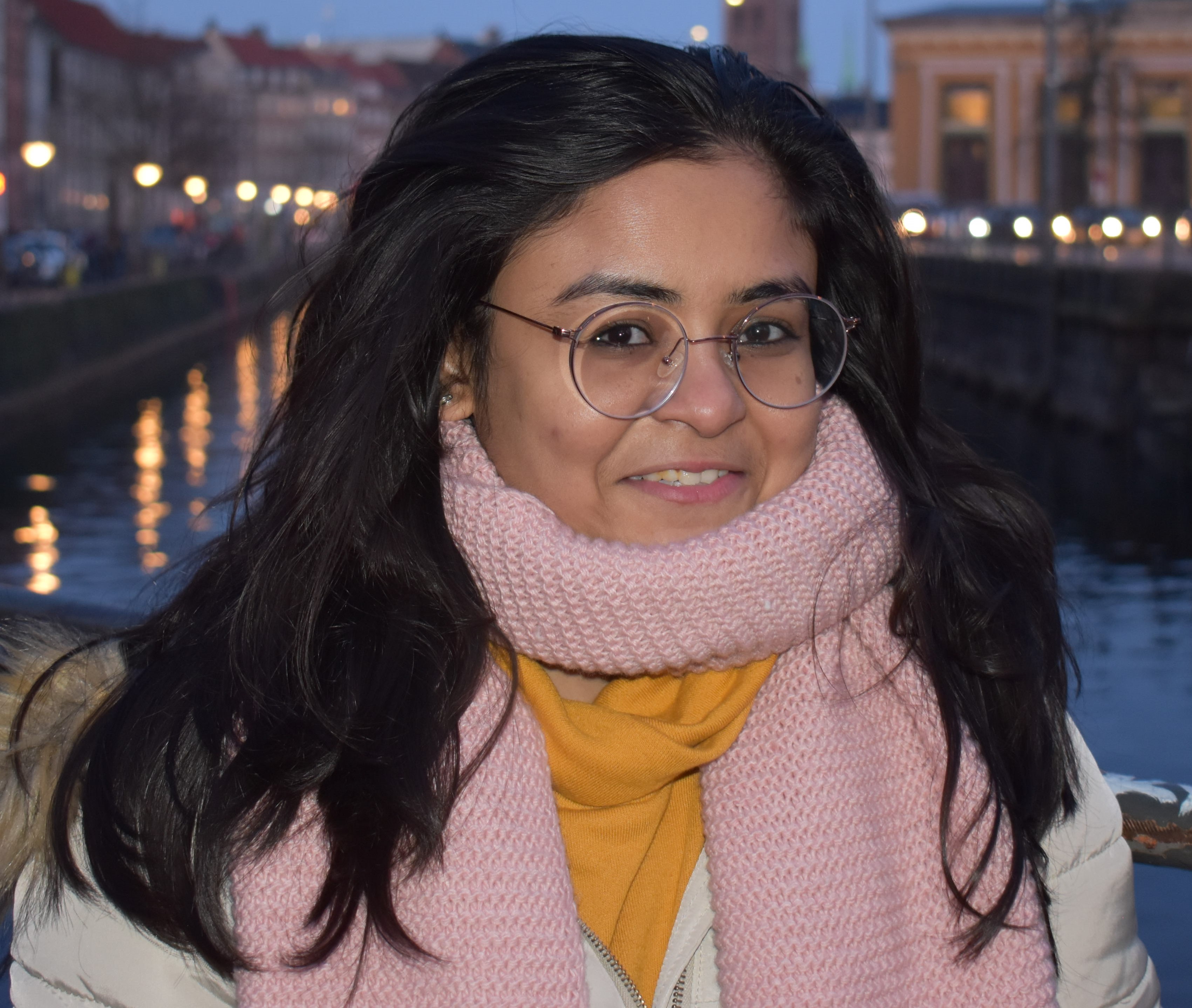}}]{Roshni Chakraborty} is an Assistant Professor at Institute of Computer Science, University of Tartu, Tartu. She was a Postdoctorate Research Fellow at Center for Data-Intensive Systems (Daisy), Aalborg University, Denmark from November 2020 to November 2022. She received her PhD degree from IIT Patna, India in 2020 and M.E. degree from IIEST Shibpur, India in 2014. Her research interests include Computational Journalism, Social Computing, Time-Series Analytics and Signed Networks.  
\end{IEEEbiography}

\begin{IEEEbiography}[{\includegraphics[width=1in,height=1.25in,clip,keepaspectratio]{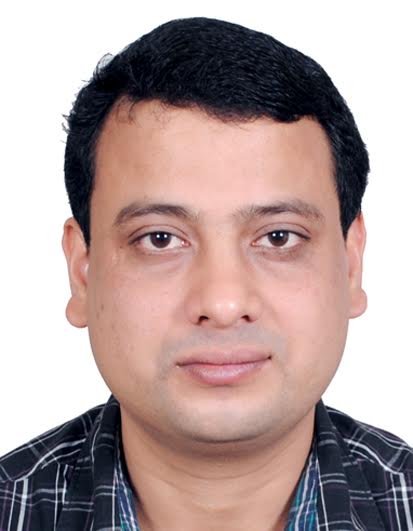}}]{Sourav Kumar Dandapat}
is an Assistant Professor of Indian Institute of Technology Patna from 2016, February onward. He completed his PhD in 2015 and M.Tech in 2005 from Indian Institute of Technology Kharagpur, India. He received his B.E degree from Jadvapur University, West Bengal, India in 2002. His current research interest includes Computational Journalism, Social Computing, Information Retrieval, Human-Computer Interaction, etc.
\end{IEEEbiography}

\end{document}